# Self Organization Map based Texture Feature Extraction for Efficient Medical Image Categorization

Marghny.H.Mohamed
Faculty of computers and Information
Assiut University, Assiut, Egypt.

Mohammed.M.Abdelsamea
Faculty of Science
Assiut University, Assiut, Egypt.

## ABSTRACT

Texture is one of the most important properties of visual surface that helps in discriminating one object from another or an object from background. The self-organizing map (SOM) is an excellent tool in exploratory phase of data mining. It projects its input space on prototypes of a low-dimensional regular grid that can be effectively utilized to visualize and explore properties of the data. This paper proposes an enhancement extraction method for accurate extracting features for efficient image representation it based on SOM neural network. In this approach, we apply three different partitioning approaches as a region of interested (ROI) selection methods for extracting different accurate textural features from medical image as a primary step of our extraction method. Fisherfaces feature selection is used, for selecting discriminated features form extracted textural features. Experimental result showed the high accuracy of medical image categorization with our proposed extraction method. Experiments held on Mammographic Image Analysis Society (MIAS) dataset.

## 1. INTRODUCTION

Image analysis techniques have played an important role in several medical applications. In general, the applications involve the automatic extraction of features from the image which is then used for a variety of classification tasks, such as distinguishing normal tissue from abnormal tissue. Chabat [1] used 13 texture parameters, derived from the histogram, co-occurrence matrix and run-length matrix categories, to differentiate between a variety of obstructive lung diseases in thin-section CT images. Kovalev [2] used texture parameters derived from gradient vectors and from generalized co-occurrence matrices for the characterization of texture of some MRT2 brain images. Herlidou [3] used texture parameters based on the histogram, co-occurrence matrix, gradient and run-length matrix for the characterization of healthy and pathological human brain tissues (white matter, grey matter, cerebrospinal fluid, tumours and oedema). Mahmoud [4] used the texture analysis approach based on a three-dimensional co-occurrence matrix in order to improve brain tumour characterization. Du-Yih Tsai [5] used four texture features derived from the co-occurrence matrix was used for classification of the heart disease. H.S. Sheshadri [6] used Six textural features for mammogram images derived from the histogram categories was used as a part of developing a computer aided decision system for early detection of breast cancer. Maria-Luiza [7] used texture parameters based on the histogram for tumour classification in mammograms. Feature extraction [8] is a vital component of the Computer Aided Diagnosis (CAD) System [9] that can discriminate between medical tissues to serve as a second reader to aid radiologists. The feature extraction unit is used to prepare data in a form that is easy for a decision support system or a classification unit to use. Compared to the input, the output data from the feature extraction unit is usually of a much lower dimension as well as in a much easier form to classify. Medical images possess a vast amount of texture information relevant to clinical practice [10, 11]. Hence texture is the most promising feature to work on. Texture analysis gives information about the arrangement and spatial properties of fundamental image elements [12]. Coggins [13] has compiled a catalogue of texture definitions. One of the most commonly used texture parameters come from Co-occurrence matrix as a statistical approach [10] which represents texture in an image using properties governing the distribution and relationships of grey-level values in the image methods normally achieve higher discrimination indexes than the structural or transform methods. This paper introduces a three different clustering approach as a region of interest selection (ROI) [14, 15] approach which provides different accurate extraction of textural features for efficient image feature representation. We use four texture features measured from a gray-level co-occurrence matrix generated from the breast images for classification of the images. A statistical discrimination method (fisherfaces algorithm) [16, 17] for feature selection algorithm is also used for extracting discriminative information from extracted feature of medical images to be used as inputs to our classification system.

The SOM [18, 19] has shown to be a highly effective tool for data visualization in a broad spectrum of application domains. One important characteristics of SOM is that it can simultaneously perform the feature extraction and it performs the classification as well [20]. Also its main property, called topology preservation, is that two individuals classified in neighboring classes are close in the input space. Even so, the complexity of the data can be challenging and leads to several interpretations.

In these paper we set up our SOM to transform an incoming signal pattern (feature set of images) of arbitrary dimension into a two dimensional lattice by creates a set of Prototype vectors representing the feature set. Our proposed features come from prototypes of trained SOM neural network, in the experiment, these new features are tested in several way and with several extraction method, which proved that it yield highest performance in overall accuracy of classification system.

## 2. DATA COLLECTION AND PRE-PROCESSING PHASE

The data collection which has been used in our experiments was taken from the MIAS [21]. The same collection has been used in other studies of automatic mammography classification. Its corpus consists of 322 images, which belong to three big categories: normal, benign and malign. There are 208 normal images, 63 benign and 51 malign, which are considered abnormal. In addition, the abnormal cases are further divided in six categories: microcalcification, circumscribed masses, speculated masses, ill-defined masses, architectural distortion and asymmetry.

Mammograms are images difficult to interpret, and a preprocessing phase of the images is necessary to improve the quality of the images and make the feature extraction phase more reliable. We applied to the images two techniques: a cropping operation and an image enhancement one. The first one was employed in order to cut the black parts of the image as well as the existing. Image enhancement helps in qualitative improvement of the image with respect to a specific application. In order to diminish the effect of over brightness or over darkness in the images and accentuate the image features, we applied a widely used technique in image enhancement to improve visual appearance of images known as Histogram Equalization. This process equalizes the illumination of the image and accentuates the features to be extracted [7].

## 3. FEATURE EXTRACTION

A major component in analyzing images involves data reduction which is accomplished by intelligently modifying the image from the lowest level of pixel data into higher level representations. From these higher level representations we can gather useful information; a process called feature extraction [8]. Extracting features by fixed blocs in the image has been considered to be sufficient as an ROI selection method in some medical applications where a large fraction of the image is covered by tissue of interest.

### 3.1. Measurements of Texture Features

The gray-level co-occurrence matrix (GLCM) is a matrix used to express the correlation of spatial location and gray-level distribution of an image [22]. From it, the local variation of gray levels on an image or sub-image can be statistically investigated and in turn, enable us to know the manner of change in gray level as a whole. In the current application, we used the following conditions to generate gray-level co-occurrence matrices.

- Direction: In general the gray-level co-occurrence matrices from 0, 45, 90, and 135 directions are used. Only the direction of 0 was used in the study.
- Distance: The length of 1-pixel was used.

Of the 14 original statistics developed by Haralick et al. (1973) for generating texture features based on co-occurrence probabilities, we chose the four most commonly used features for our evaluation.

Dissimilarity, uniformity, entropy, and contrast are often used in practice. The following four statistics will be used exclusively in this paper:

1. $\text{Dissimilarity} = \sum_{i,j=1}^{G} C_{ij} \left| i - j \right|.$

2. $\text{Uniformity} = \sum_{i,j=1}^{G} C_{ij}^{2}.$

3. $\text{Entropy} = -\sum_{i,j=1}^{G} C_{ij} \lg C_{ij}.$

4. $\text{Contrast} = \sum_{i,j=1}^{G} C_{ij} (i - j)^{2}.$

Where, Cij represents co-occurring probabilities stored inside GLCM. G represents number of grey level available.

Now, we introduce two ROI selection methods [23] that give us better texture feature extraction when co-occurrence matrix method is used and hence a more accurate texture feature.

#### 3.1.1. Extraction based on pixel wise segmentation approach.

For a more accurate feature extraction, apply extraction based pixel wise partitioning method as follows:

**Step1:** divide the entire image into SN non-overlapping sub-images SI= {I₁,I₂,...,I_{SN}}.

**Step2:** use the k-means algorithm [24] to cluster the sub-images (SI) into several classes based on pixel intensity for each I_i, i=1,2,...,SN independently.

**Step3:** for each cluster in I_i, i=1,2,...,SN, construct a sub-image representing set of texture feature vectors F_K ={f₁,f₂,...,f_X}, k=1,2,...,L; where L is the number of classes each of which contains X texture features.

**Step4:** build the final set of texture features representing the overall image in the form of a single transaction of the final dataset (set of images), T={t₁,t₂,...,t_c}, where c is the number of images, t_i is a vector of the size (SN × L × X), i=1, 2, ..., c.

**Step5:** for each t_i, i=1,2,...,c add the class label of its image.

#### 3.1.2. Extraction method based on bloc wise segmentation approach.

For a more accurate feature extraction and a further investigation of the localization, apply extraction based bloc wise partitioning method as follows:

**Step1:** divide the entire image into SN non-overlapping sub-images SI= {I₁,I₂,...,I_{SN}}.

**Step2:** split each of these SN sub-image into other M blocs I_j={B₁,B₂,...B_M}, j=1,2,...,SN.

**Step3:** for each bloc B_i, i=1,2,...,M, construct a bloc representing set of texture feature vectors.

**Step4:** use the k-means algorithm to cluster the feature vectors into several classes for each sub-image I independently.

**Step5:** for each cluster in I_L i=1,2,...,SN, construct a sub-image representing set of texture feature vectors F_K ={f₁,f₂,...,f_X}, k=1,2,...,L; where L is the number of classes each of which contains X texture features.

**Step6:** build the final set of texture features representing the overall image in the form of a single transaction of the final dataset (set of images), T={t₁,t₂,...,t_c}, where c is the number of images, t_i is a vector of the size (SN × L × X), i=1, 2, ..., c.

**Step7:** for each t_i, i=1,2,...,c add the class label of its image.

### 3.2. Enhancement extraction method based on SOM

Our SOM consists of a regular, two dimensional (2-D), grid of map units. Each unit I is represented by a prototype vector WI = {WI1 ,WI2 , ...,WIS}, where S is input vector dimension. The units are connected to adjacent ones by a neighborhood relation. Our proposed features for improved final reduced textural features come from prototypes of SOM network rather than textural features as a new features based on old texture features for efficient image representation. In the experiment, we compared our old textural feature that extracted from the three previous extraction methods and selected by fisherfaces method with its new features based SOM. Our proposed method described as:

**Step1**: After constructing texture vectors for all images from any of previous three extraction method (TOld) a fisherfaces feature selection is used as image feature data set adaptation for dimensionality reduction to get final reduced textural feature, called textural features set, finally set an old textural feature set in the form of transaction TNew = {t1, t2, . . . , tc}, where c is the number of images, every t is a vector of the size S in addition to the class label of the image.

**Step 2:** Randomly initialize the weight vector of SOM map.

**Step 3:** Set iteration = 0.

**Step 4:** Select a single transaction of the old textural feature set, TNew = {t1, t2, . . . , tc}.

**Step 5:** Find a winner neuron, the best matching unit (BMU), on the SOM for that input data using similarity Measure

$$\left\| X - W_I \right\| = \min_I \left[ \left\| X - W_I \right\| \right]$$

**Step 6:** Set the learning rate and the neighborhood function according to iteration number and update the winner neuron and its neighbors on the SOM map as:

$$W_I(t+1) = W_I(t) + \alpha(t)h_{WI}(t)[X - W_I(t)],$$

where

t : time.

$\alpha(t)$ : Adaptation coefficient.

hWI (t) : Neighborhood kernel centered on the winner unit:

$$h_{WI}(t) = \exp\left(-\frac{\left\| r_w - r_I \right\|^2}{2\sigma^2(t)}\right),$$

where

$r_w$ and $r_I$ are positions of neurons W and I on the

SOM grid. Both $\alpha(t)$ and $\sigma(t)$ decrease monotonically with time.

**Step 7:** Set Iteration = Iteration + 1.

**Step 8:** If Iteration reaches its maximum value go to next step, otherwise go to Step 3.

**Step 9:** Find the final winner neuron for a single transaction of the old data set and replace feature vector of that transaction with the prototype of its winner neuron as a new feature vector for represent that image. In the experimental we checked the the possibility of adding these proposed features to textural feature set rather than replacing (see Table.4).

**Step 10:** Repeat Step 9 for all transaction of the old data set.

**Step 11:** Finally, set a new features based SOM data set in the form of transaction as the input data set TSOM = {t1, t2, . . . , tc}, where c is the number of images, every t is a vector of the size (S) in addition to the class label of the image.

Proposed extraction method illustrated in fig 1.

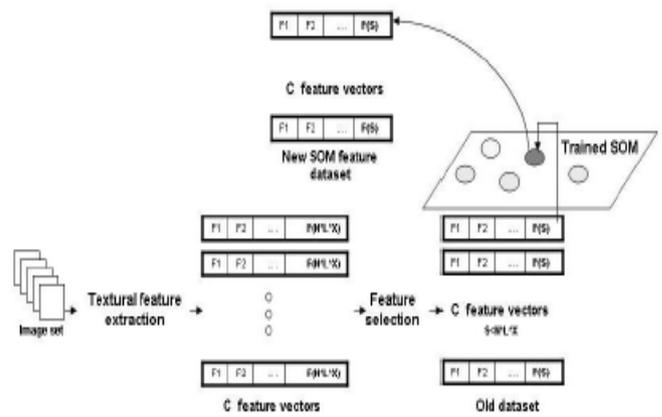

Fig.3 Generated features based SOM.

## 4. FEATURE SELECTION

The statistical discrimination methods are suitable not only for classification but also for characterization of differences between a reference group of patterns and the population under investigation. For the image classification case, without class labels, Principle Component Analysis (PCA) can be applied, when some class labels are available, linear discriminant analysis (LDA) is applied to transform the features into the most discriminating Feature space [25-29]. Let us consider a set of N sample images {x1,x2,…,xN} taking values in an n-dimensional image space, and assume that each image belongs to one of c classes. Let us also consider a linear transformation mapping the original n-dimensional image space into an m-dimensional feature space, where m < n. The new feature vectors Y are defined by the following linear transformation:

$$Y_k = W^T x_k, \qquad\qquad k=1,N. \qquad (1)$$

### 4.1. Fisher's Linear Discriminant Analysis (FLD)

In this method, W in (1) can be selected in such a way that the ratio of the between-class scatter and the within-class scatter is maximized. Let the between-class scatter matrix be defined as

$$S_b = \sum_{i=1}^{c} N_i (\mu_i - \mu)(\mu_i - \mu)^T,$$

and the within-class scatter matrix be defined as

$$S_w = \sum_{i=1}^{c} \sum_{x_k \in X_i} (x_k - \mu_i)(x_k - \mu_i)^T,$$

where $\mu_i$ is the mean image of class $X_i$, and $N_i$ is the number of samples in class $X_i$. If $S_w$ is nonsingular, the optimal projection $W_{opt}$ is chosen as the matrix with orthonormal columns which maximizes the ratio of the determinant of the between-class scatter matrix of the projected samples to the determinant of the within-class scatter matrix of the projected samples, i.e.,

$$W_{opt} = \arg\max_{W} \frac{\left| W^{T} S_{b} W \right|}{\left| W^{T} S_{w} W \right|}. \qquad (2)$$

In this paper, we can not apply (FLD) directly to solve the recognition problem since the dimension of the sample space is typically larger than the number of samples in the training set. As a consequence, $S_w$ is singular in this case.

### 4.1.1. Fisherfaces.

Swets and Weng [30] proposed a two stage PCA+LDA method, also known as the fisherface method, in which PCA is first used for dimension reduction so as to make Sw nonsingular before the application of LDA. In this method applying an alternative to the criterion in (2).

This method avoids this problem by projecting the image data set to a lower dimensional space so that the resulting within-class scatter matrix Sw is nonsingular. This is achieved by using PCA to reduce the dimension of the feature space to N-c, and then applying the standard FLD defined by (2) to reduce the dimension to c-1[16]. More formally, Wopt is given by

$$W_{opt}^{T} = W_{fld}^{T} W_{pca}^{T},$$

where

$$W_{pca} = \arg\max_{W} \left| W^{T} S_{T} W \right|,$$

$$W_{fld} = \arg\max_{W} \frac{\left| W^{T} W_{pca}^{T} S_{b} W_{pca} W \right|}{\left| W^{T} W_{pca}^{T} S_{w} W_{pca} W \right|},$$

$$S_{T} = \sum_{k=1}^{N} (x_{k} - \mu)(x_{k} - \mu)^{T}.$$

Note that the optimization for $W_{opt}$ is performed over $n \times (N-c)$ matrices with orthonormal columns, while the optimization for $W_{fld}$ is performed over $(N-c) \times m$ matrices with orthonormal columns. In computing $W_{opt}$, we have thrown away only the smallest (c-1) principal components.

## 5. MEASURES FOR PERFORMANCE EVALUATION

We evaluated the performance of the proposed methods in terms of sensitivity, specificity and overall accuracy [5], Such that k-fold cross validation process is used. K-fold cross validation is used for model testing and evaluation to determine how accurately a learning algorithm will be able to predict data that it was not trained on. In this method it is not important how the data is divided. Every data point appears in a test set exactly once, and appears in a training set k - 1 times. We can independently choose the size of the each test and the number of trials.

## 6. Experimental results

In our experiment a sample of 22% from the MIAS dataset is selected randomly for model testing and evaluation. The sample is distributed between class as following: Normal class (n=30), microcalcification class (n=10), circumscribed masses class (n=7), speculated masses class (n=8), ill-defined masses class (n=5), architectural distortion class (n=7) and asymmetry class (n=4).
We evaluate the performance of the proposed extraction method in terms of overall accuracy with standard classifiers, using Weka experimenter [31], Such that partitioning our data set using 10-fold cross-validation. Weka is a collection of machine learning algorithms for data mining tasks. The algorithms can either be applied directly to a dataset or called from your own Java code. Weka contains tools for data pre-processing, classification, regression, clustering, association rules, and visualization. It is also well-suited for developing new machine learning schemes.

Table.1 illustrates overall accuracy comparison between textural feature set based on fixed bloc segmentation (simply divide the whole image into 6 non overlapping sub-images then split each of these 6 sub images into other 8 non-overlapping blocs). and its enhancement feature set extracted from proposed method (Several values of SOM map size were checked for the possibility of improving the results), while feature extraction in regardless to feature selection and classification algorithms used, See table 1.

Table.1 the classification rate comparison.

| Classifiers | FixedBloc Features | Features BasedSOM with MapSize | | |
|---|---|---|---|---|
| | | 5x5 | 10x10 | 15x15 |
| NNGe | 100% | 91.54% | 100% | 100% |
| J48 | 90.14% | 91.54% | 95.77% | 95.77% |
| Bayes Net | 98.59% | 91.54% | 100% | 100% |
| NaiveBayes | 100% | 91.54% | 100% | 100% |
| Bagging | 95.77% | 92.95% | 100% | 95.77% |
| RBF Net | 100% | 91.54% | 100% | 100% |

from results shown in the table.1, we can notice that, our proposed method gave better accuracy than extraction method based on fixed bloc partitioning as a ROI selection method. See fig.2.

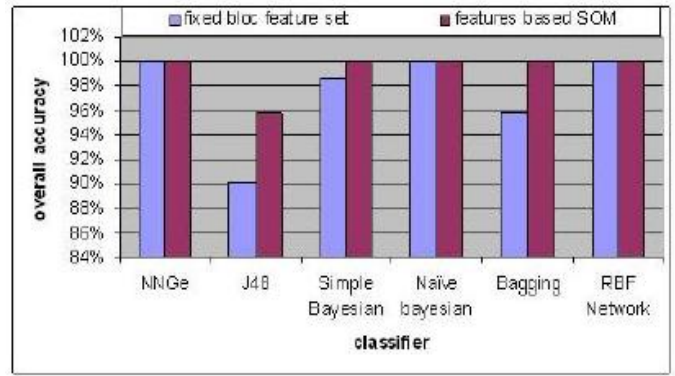

Fig.2 Success rate of classifiers with our proposed method.

Table.2 present comparison between textural feature sets based on pixel intensity segmentation described in section 3.1.1 with SN=6 and L=3 and its enhancement feature set extracted from proposed method.

Table.2 the classification rate comparison.

| Classifiers | PixelWise Features | Features BasedSOM with MapSize | | |
|---|---|---|---|---|
| | | 5x5 | 10x10 | 15x15 |
| NNGe | 100% | 88.73% | 100% | 100% |
| J48 | 92.95% | 90.14% | 97.18% | 92.95% |
| Bayes Net | 98.59% | 91.54% | 100% | 100% |
| NaiveBayes | 100% | 91.54% | 100% | 98.59% |
| Bagging | 100% | 88.73% | 100% | 98.59% |
| RBF Net | 100% | 83.09% | 100% | 100% |

from results shown in the table.2, we can notice that, our proposed method gave better accuracy than extraction method based on pixel wise partitioning as a ROI selection method. See fig.3.

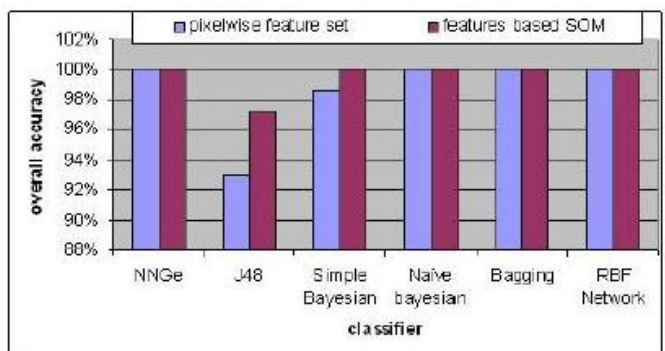

Fig.3 Success rate of classifiers with our proposed method.

Table.3 present comparison between textural feature sets based on our bloc wise segmentation described in section 3.1.2 with SN=6, M=8 and L=3 and its enhancement feature set extracted from proposed method.

Table.3 the classification rate comparison.

| Classifiers | BlocWise Features | Features BasedSOM with MapSize | | |
|---|---|---|---|---|
| | | 5x5 | 10x10 | 15x15 |
| NNGe | 100% | 98.59% | 100% | 98.59% |
| J48 | 94.36% | 88.73% | 97.18% | 95.77% |
| Bayes Net | 100% | 88.73% | 100% | 100% |
| NaiveBayes | 100% | 88.73% | 100% | 98.59% |
| Bagging | 98.59% | 92.95% | 100% | 98.59% |
| RBF Net | 100% | 88.73% | 100% | 100% |

from results shown in the table.3, we can notice that, our proposed method gave better accuracy than extraction method based on bloc wise partitioning as a ROI selection method. See fig.4.

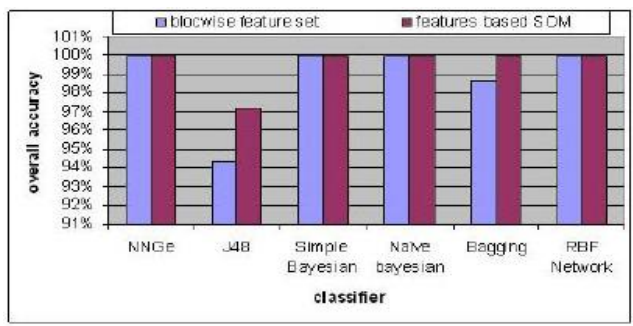

Fig.4 Success rate of classifiers with our proposed method.

Now we illustrates the possibility of improving result when adding enhancement textural features extracted from proposed extraction method with its relevant textural feature set for more accurate image textural representation.
Table.2 illustrates comparison between the following three textural feature sets based on different ROI selection methods and there relevant enhancement feature sets extracted from proposed extraction method with $10 \times 10$ map size:

1. Textural feature sets based on fixed bloc segmentation (simply divide the whole image into 6 non overlapping sub-images then split each of these 6 sub-images into other 8 non-overlapping blocs).

2. Textural feature sets based on pixel intensity segmentation described in section 3.1.1 with SN=6 and L=3.

3. Textural feature sets based on our bloc wise segmentation described in section 3.1.2 with SN=6, M=8 and L=3

Table.4 overall accuracy comparisons.

| Feature Set Of | NNGe | J48 | bayes Net | Naive bayes | Bagging | RBF Net |
|---|---|---|---|---|---|---|
| FixedBloc | 100% | 90.14% | 98.95% | 100% | 95.77% | 100% |
| FixedBloc+ SOMBased FixedBloc | 100% | 95.77% | 100% | 100% | 97.18% | 100% |
| BixelWise | 100% | 92.95% | 98.59% | 100% | 100% | 100% |
| BixelWise+ SOMBased BixelWise | 100% | 94.36% | 100% | 100% | 100% | 100% |
| BlocWise | 100% | 94.36% | 100% | 100% | 98.59% | 100% |
| BlocWise+ SOMBased BlocWise | 98.59% | 94.36% | 100% | 100% | 98.59% | 100% |

From results shown in the table.4, we can notice that, our proposed method stable and improving image representation when adding its features to its original features. See fig.5.

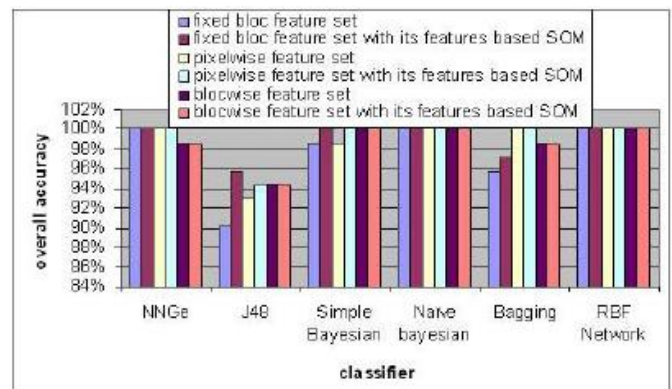

Fig.5 Success rate of classifiers with modification of our proposed method.

## 7. CONCLUSIONS.

In this work, four texture features derived from the co occurrence matrix was used as a part of developing CAD system for early detection of breast cancer. This paper present textural extraction method based on three different ROI selection methods for obtaining efficient image representation. A fisherfaces feature selection algorithm is used for extracting discriminative information from extracted feature of medical images.

This paper present an enhancement method based on prototypes of SOM network for accurate extraction of textural feature set. Our experiments confirms that proposed technique improves the localization of the texture feature, produce general and accurate set of features from images for efficient image representation contents, and improves the final decision of the CAD. From our experiments, the selected classifiers RBF network showed the highest performance when compared to others, as RBF network is very much suitable for such application.

## 8. REFERENCE.